%% file: egpaper_final.tex
\documentclass[10pt,twocolumn,letterpaper]{article}

\usepackage{cvpr}
\usepackage{times}
\usepackage{epsfig}
\usepackage{graphicx}
\usepackage{amsmath}
\usepackage{amssymb}
\usepackage{booktabs}

\usepackage[pagebackref=true,breaklinks=true,letterpaper=true,colorlinks,bookmarks=false]{hyperref}

\cvprfinalcopy 

\def\cvprPaperID{****} 

\begin{document}

\title{Keep it Simple: Image Statistics Matching for Domain Adaptation}

\author{Alexey Abramov$^{1}$ \thanks{Equal contribution.}
\and
Christopher Bayer \footnotemark[1] \space \thanks{This work was done while at Continental Teves AG.}
\and
Claudio Heller$^{1}$ \footnotemark[1]\\
$^{1}$Continental Teves AG, Frankfurt am Main, Germany\\
{\tt\small \{alexey.abramov,claudio.heller\}@continental.com, c4.bayer@gmail.com}
}

\maketitle

\input{content.tex}

{\small
\bibliographystyle{ieeetr}
\bibliography{egpaper_final}
}

\end{document}

%% file: content.tex
\begin{abstract}
\input{abstract.tex}
\end{abstract}

\section{Introduction}
\input{introduction.tex}

\section{Related Work}
\label{chap:related-work}
\input{related_work.tex}

\section{Methodology}
\label{chap:methods}
\input{methods.tex}

\section{Experimental Evaluation}
\label{chap:results}
\input{experimental_results.tex}

\section{Conclusion}
\label{chap:conclusion}
\input{conclusion.tex}


%% file: abstract.tex
Applying an object detector, which is neither trained nor fine-tuned on data close to the final application, often leads to a substantial performance drop. In order to overcome this problem, it is necessary to consider a shift between source and target domains. Tackling the shift is known as \textit{Domain Adaptation (DA)}. In this work, we focus on unsupervised DA: maintaining the detection accuracy across different data distributions, when only unlabeled images are available of the target domain. Recent state-of-the-art methods try to reduce the domain gap using an adversarial training strategy which increases the performance but at the same time the complexity of the training procedure. In contrast, we look at the problem from a new perspective and keep it simple by solely matching image statistics between source and target domain. We propose to align either color histograms or mean and covariance of the source images towards the target domain. Hence, DA is accomplished without architectural add-ons and additional hyper-parameters. The benefit of the approaches is demonstrated by evaluating different domain shift scenarios on public data sets. In comparison to recent methods, we achieve state-of-the-art performance using a much simpler procedure for the training. Additionally, we show that applying our techniques significantly reduces the amount of synthetic data needed to learn a general model and thus increases the value of simulation.

%% file: introduction.tex
\begin{figure}[t]
	\begin{center}
		\includegraphics[width=8.3cm]{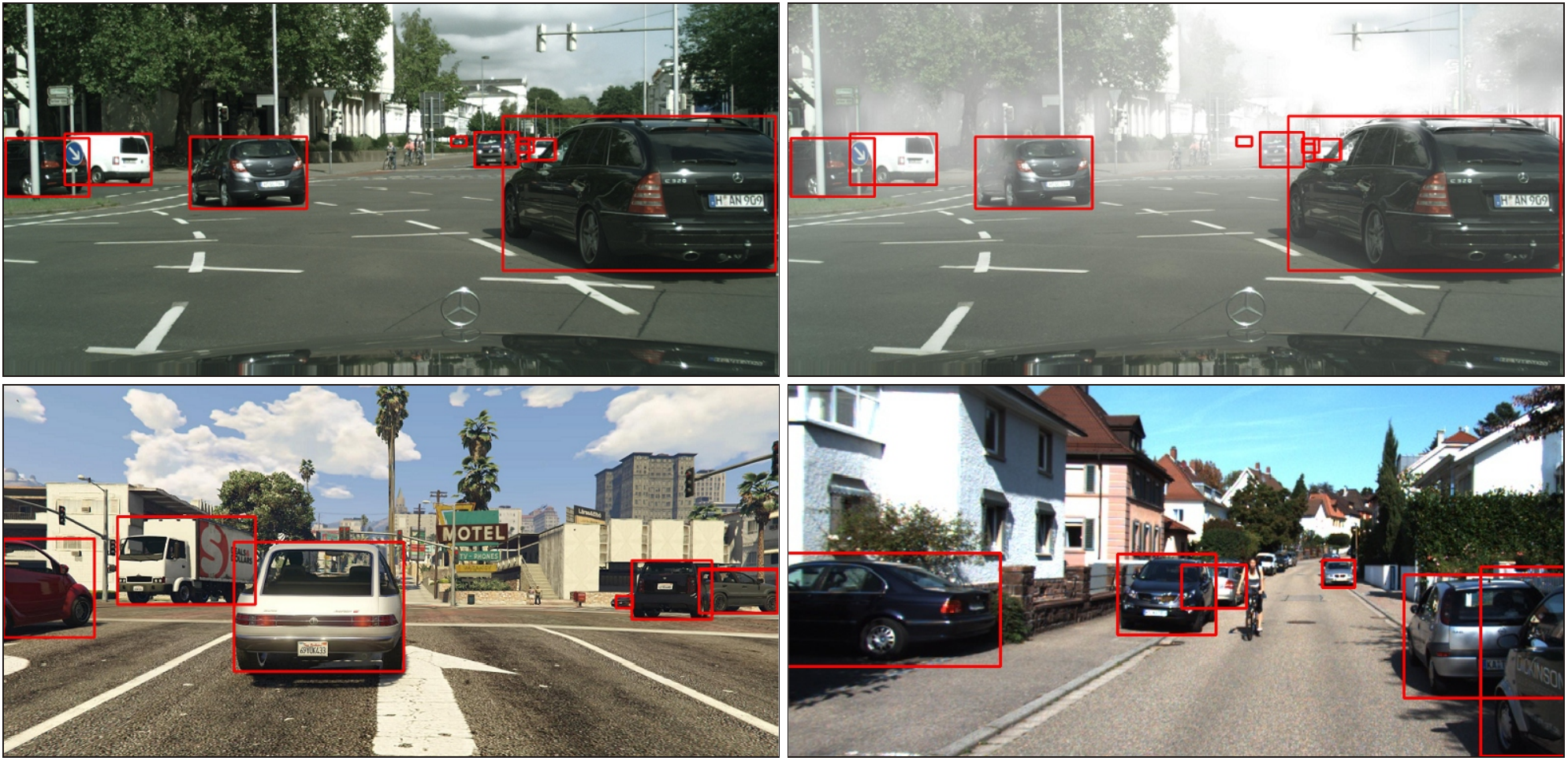}
		\caption{Ground truth bounding boxes for the class \textit{car} on different public benchmarks with urban traffic scenes. From left to right and from top to bottom: \textit{Cityscapes}~\cite{Cordts2016}, \textit{Foggy Cityscapes}~\cite{Sakaridis2018}, \textit{GTA~Sim~10k}~\cite{Roberson2017}, \textit{KITTI}~\cite{Geiger2012}.
			\label{fig:benchmarks-bboxes}}
	\end{center}
\end{figure}

State-of-the-art object detectors perform very well when training and test data are drawn from the same distribution~\cite{Ren2015, Redmon2017, Liu2016}. However, in practice there is often a mismatch between source (training) and target (testing) domains. Considerable differences in lighting, viewpoints, image quality, object appearances, contexts, etc. cause a significant performance drop due to data set bias, when models, trained in one domain, are naively applied to other domains~\cite{Gopalan2011,Herranz2016,Torralba2011}. \textit{Domain Adaptation (DA)} approaches aim at maintaining detection accuracy in the presence of a shift between distributions~\cite{BenDavid2010,Tzeng2015}. In this paper, we focus on \textit{unsupervised DA}: images and annotations are available in the source domain (full supervision), while only unlabeled images are available for the target domain.

Driving scenarios are especially challenging, since camera setups, used for collecting training data, might differ from those in particular cars. Besides that, the vehicles might operate in diverse environments, where infrastructure and objects look slightly or notably different. Furthermore, due to an immense variety in urban street scenes it is physically impossible to collect a training data set, which is large and comprehensive enough for learning a general model. Unusual, rare or emergency scenarios can be very difficult or even dangerous to acquire. In order to overcome the issue of finite labeled data sets, the use of photo-realistic computer simulation became recently very important for generating synthetic images with corresponding annotations~\cite{Roberson2017, Richter2016, Ros2016, Gaidon2016}. However, due to the still present mismatch between synthetic and real data, beneficially using data from a simulation engine poses a challenge for applying a model learned in one domain to another. 

The problem of \textit{unsupervised DA} for object detection has been addressed in the past few years. Earlier methods propose an adaptive SVM~\cite{Xu2014}, subspace alignment for source and target data~\cite{Raj2015}, and learning detectors from alternative sources~\cite{Hattori2015}. More recent approaches extend a detection model by additional adaptation components employing an adversarial training strategy for learning domain-invariant features~\cite{Chen2018,Saito2019}. Alternatively, several unsupervised models based on \textit{Generative Adversarial Networks (GANs)}~\cite{Goodfellow2014} have been proposed for cross-domain image-to-image translation~\footnote{This class of methods is also known as \textit{unpaired pixel-level DA}.}~\cite{Kim2017,Liu2017,Yi2017,Taigman2017,Zhu2017}. Such techniques can be used to generate images that look like those in the target domain from images in the source domain with available annotations~\cite{Inoue2018}.

Despite the considerable progress in \textit{unsupervised DA} for object detection, much less attention has been given to analysis of discrepancies between source and target images in terms of global image characteristics. In this work, we will show that alignment of color statistics and histograms is unexpectedly powerful for reducing the domain gap. We propose to use a generalized color transfer~\cite{xiao2006color} approach for aligning mean and covariance of color channels or high-dimensional features, which we call \textit{Feature Distribution Matching} (see section~\ref{sec:fdm}). We also show that \textit{Histogram Matching} operation leads to similar results (see section~\ref{sec:hm}). Both techniques solely modify source images used for the training, while keeping the object detection model itself unchanged. Thus, no adversarial training manner is needed and there are no additional hyper-parameters. To demonstrate the effectiveness of our procedure, we perform a variety of domain shifts between well-known public benchmarks: \textit{GTA Sim 10k}~\cite{Roberson2017}, \textit{KITTI}~\cite{Geiger2012}, \textit{Cityscapes}~\cite{Cordts2016}, and \textit{Foggy Cityscapes}~\cite{Sakaridis2018} (see fig.~\ref{fig:benchmarks-bboxes}). Additionally, we show that transforming source images using the \textit{Feature Distribution Matching} and \textit{Histogram Matching} reduces amounts of training data needed to achieve good performance on the target domain.

The structure of the paper is as follows. In section~\ref{chap:related-work}, we give an overview of existing methods. In section~\ref{chap:methods}, we present the \textit{Feature Distribution Matching} and \textit{Histogram Matching} techniques in the context of \textit{unsupervised DA} for object detection. In section~\ref{chap:results}, we show the results of a versatile experimental evaluation on the public data sets, and, in section~\ref{chap:conclusion}, we discuss the results and conclude.

%% file: related_work.tex
The problem of \textit{unsupervised DA} has been addressed over the past years for the most active topics of research in computer vision, such as image classification~\cite{Fernando2013,Raj2015,Ganin2015,Ghifary2016,Li2016,Sener2016,SunSaenko2016,TzengHoffman2017,Busto2017}, semantic segmentation~\cite{Hoffman2016,Hoffman2017,Zhang2017,Huang2018} and object detection~\cite{Inoue2018,Chen2018,Saito2019}. In this work, we focus on the object detection task,
which has the goal to predict both category and location in the form of a bounding box (see fig.~\ref{fig:benchmarks-bboxes}).

To our knowledge there are only two approaches that address \textit{unsupervised DA} for the object detection task in driving scenarios~\cite{Chen2018,Saito2019}. Both methods are based on the state-of-the-art \textit{Faster R-CNN} model~\cite{Ren2015}. In \textit{Domain Adaptive Faster R-CNN} (\textit{DA Faster R-CNN}), the original model is extended by two \textit{DA} components to overcome the domain discrepancy on image and instance level~\cite{Chen2018}. Beyond that, a consistency regularization between the domain classifiers on different levels is incorporated in order to learn a domain invariant region proposal network. In each component a domain classifier is trained in an adversarial training manner using a \textit{Gradient Reverse Layer (GRL)}~\cite{Ganin2015}.

However, aligning features at the global image level may fail for object detection, since domains could have very different backgrounds, scene layouts, the number and scale of objects. Motivated by this evidence, \textit{Strong-Weak Domain Alignment} (\textit{Strong-Weak DA}) model proposes a combination of \textit{weak global} and \textit{strong local} alignments~\cite{Saito2019}. The weak global alignment model regulates an adversarial alignment loss towards images that are globally similar and away from images that are globally dissimilar. Strong local alignment is designed in a way that it only considers local features and ensures the style alignment of images across domains (color, texture). Both \textit{DA Faster R-CNN} and \textit{Strong-Weak DA} integrate additional components into the \textit{Faster R-CNN}, which employ an adversarial alignment loss using the \textit{GRL}. This introduces additional hyper-parameters to the system: both methods use a so-called \textit{trade-off parameter} to balance the \textit{Faster R-CNN} loss and the added \textit{DA} components, whereas \textit{Strong-Weak DA} needs one more parameter to control how strictly features between domains are aligned. These parameters need to be adjusted for every particular domain shift.

Inoue et al.~\cite{Inoue2018} take a different path and generate images that look like those in the target domain from images in the source domain. The generation is achieved by unpaired image-to-image translation using \textit{Cycle-Consistent Adversarial Network} (referred as \textit{CycleGAN})~\cite{Zhu2017}. A fully supervised detector is trained thereafter on the generated images.

Our approach, on the contrary, does not require any architectural changes, extensions, or hyper-parameters to be tuned, it only alters source input images according to those in the target domain. There is also no need in a special training schedule for the model, since input images are already transferred into the target domain and a common training process can be applied. This makes our method significantly faster at training time in comparison to \textit{DA Faster R-CNN}, \textit{Strong-Weak DA}, and \textit{CycleGAN}~\footnote{In case source input images are transformed beforehand, there is no overhead in the training time at all, as the training procedure is identical to the one of the original model.}.

\begin{figure*}[!ht]
	\begin{center}
		\includegraphics[width=17.5cm]{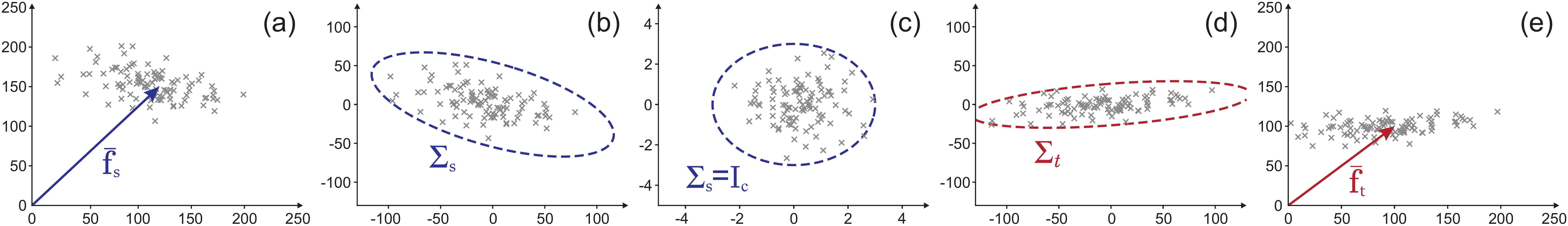}
		\caption{
			Exemplary procedure of \textit{FDM} on $2D$ data (one can also interpret it as image data in the Red-Green plane, where each feature corresponds to the values $[0,255]$ of a certain pixel):
			(a) feature points (grey) from the source domain and corresponding mean vector $\mathbf{\bar{f}}_s$  (blue),
			(b) centered source feature points and the corresponding covariance $\mathbf{\Sigma}_s$ plotted as a $3\sigma$ ellipse (dashed blue),
			(c) source data and covariance ellipse after the whitening transformation of eq. \ref{whitning},
			(d) transformed features $\mathbf{F}^0_{s \rightarrow t}$ after using the target covariance $\mathbf{\Sigma}_t$ (dashed red) in eq.~\ref{recoloring},
			(e) Final transformed features $\mathbf{F}_{s \rightarrow t}$ shifted by the target mean (red).
			\label{fig:FDM}
		}
	\end{center}
\end{figure*}

%% file: methods.tex
Let $\mathcal{D}_s = \{\mathbf{x}_s^i, y_s^i\}_{i=1,...,n_s}$ and $\mathcal{D}_t = \{\mathbf{x}_t^j, y_t^j\}_{j=1,...,n_t}$ be a source and target domain, respectively, where $\mathbf{x}_s^i$, $\mathbf{x}_t^j$ are the input data samples, and $y_s^i$, $y_t^j$ the corresponding labels. 
Training a \textit{Convolutional Neural Network (CNN)} $y =\varPhi(\mathbf{x})$ on $\mathcal{D}_s$ and evaluating it afterward on $\mathcal{D}_t$ can lead to poor results, in case there is a shift between both distributions.
\textit{DA} aims to close this gap by modifying the source data in such a way that a \textit{CNN} trained on the transformed data performs better on the target domain. 
Note that for \textit{unsupervised DA}, the target labels $y_t$ are not utilized.

We present two methods for pairwise \textit{DA}, meaning that for each source image one target image is (e.g. randomly) chosen for the transformation. 
Concretely, given an image of height $h_s$ and width $w_s$ with $c$ channels (in this work $c=3$ for an RGB space) from the source domain 
$\mathbf{x}^i_s \in \mathbb{N}^{h_s \times w_s \times c}$ and an image from the target domain
$\mathbf{x}^j_t \in \mathbb{N}^{h_t \times w_t \times c}$, pairwise \textit{DA} can be written as the transformation function:
$\mathbf{x}^i_{s \rightarrow t} = \Psi(\mathbf{x}^i_s | \mathbf{x}^j_t) \in \mathbb{N}^{h_s \times w_s \times c}$.


\subsection{Feature Distribution Matching}
\label{sec:fdm}

\input{svd_swap.tex}

\subsection{Histogram Matching}
\label{sec:hm}
\input{hist_matching.tex}


%% file: svd_swap.tex
\begin{figure*}[!ht]
	\begin{center}
		\includegraphics[width=17.5cm]{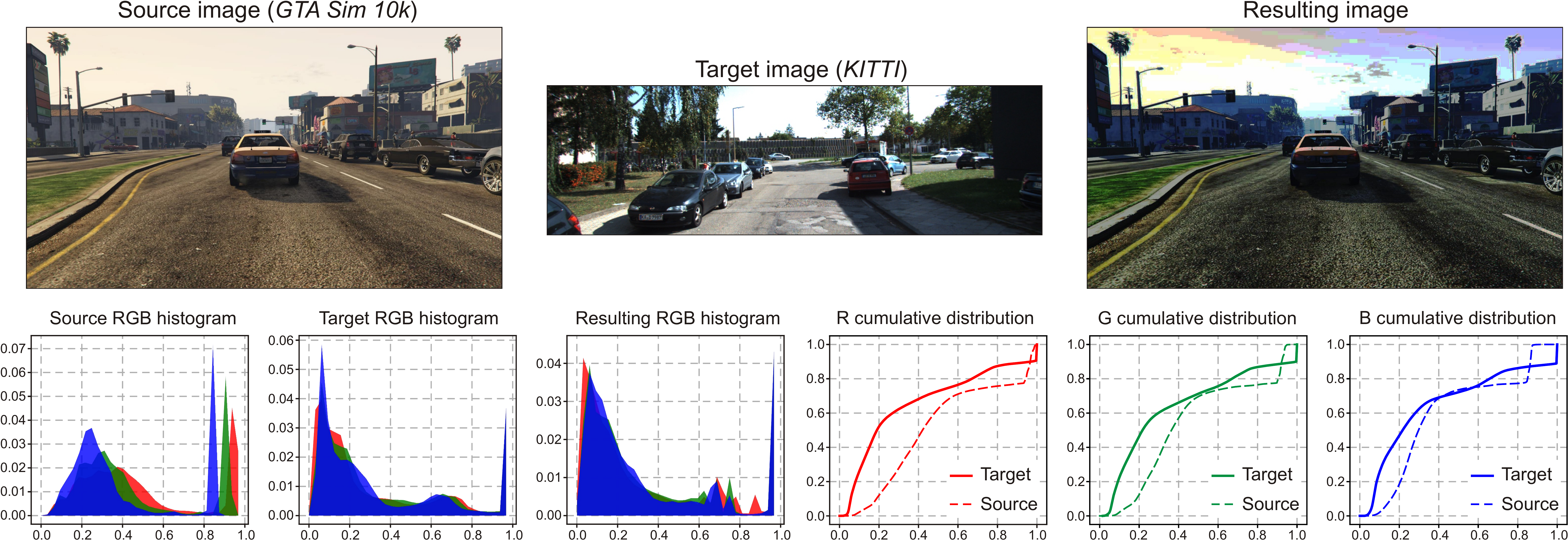}
		\caption{Histogram Matching (HM) for a pair of images: the first row shows the source image from the \textit{GTA Sim 10k} data set, the target image from the \textit{KITTI} data set and the resulting image, whereas the second row shows the corresponding color histograms along with the cumulative distribution functions for each color channel.
			\label{fig:histogram-matching}}
	\end{center}
\end{figure*}

Following the color transfer method of Xiao and Ma~\cite{xiao2006color}, the idea is to transform a source image $\mathbf{x}^i_{s}$ in such a way that it obtains the color mean and covariance of the target image $\mathbf{x}^j_{t}$, while retaining the source image content.
Instead of a transformation in homogeneous coordinates, we propose \textit{Feature Distribution Matching (FDM)}, which generalizes the transformation to the $c$-dimensional Euclidean space.

As the first step, the source and target image are reshaped into a feature matrix
\[ 
\mathbf{x}^i_s \in \mathbb{N}^{h_s \times w_s \times c}
\rightarrow
\mathbf{F}_s \in \mathbb{R}^{N_s \times c}
\]
\begin{equation}\label{oberservation_vectors}
\mathbf{x}^j_t \in \mathbb{N}^{h_t \times w_t \times c}
\rightarrow 
\mathbf{F}_t \in \mathbb{R}^{N_t \times c},
\end{equation}
where each row $\mathbf{f}^i \in \mathbf{F}$ is one of the $N$ pixels of the image and corresponds to a sample feature in the $c$ dimensional space. 
In the second step, we center the data by subtracting the sample mean:
\[ 
\mathbf{F}^0_s = \mathbf{F}_s - \mathbf{\bar{f}}_s
\]
\begin{equation}\label{center_data}
\mathbf{F}^0_t = \mathbf{F}_t - \mathbf{\bar{f}}_t, 
\end{equation}
with $\mathbf{\bar{f}} = \frac{1}{N} \sum_N \mathbf{f}^i \in \mathbb{R}^{1 \times c}$.
Next, PCA-Whitening transformation~\cite{kessy2018optimal} is applied on the source data  
by using \textit{Singular Value Decomposition (SVD)} of the covariance matrix on the centered source feature matrix.
The resulting matrix $\mathbf{U}$ is then used to rotate, and the diagonal matrix of the eigenvalues $\mathbf{S}$ to scale the sample points
\[ 
\mathbf{\Sigma}_s = cov(\mathbf{F}^0_s) \in \mathbb{R}^{c \times c}
\]
\[ 
\mathbf{U}_s \mathbf{S}_s \mathbf{V}^*_s = svd(\mathbf{\Sigma}_s)
\]
\begin{equation}\label{whitning}
\mathbf{\hat{F}}^0_s = \mathbf{F}^0_s  \mathbf{U}_s  \mathbf{S}^{-\frac{1}{2}}_s,
\end{equation}
given $cov(\mathbf{F}^0_s) = \frac{1}{n-1} \sum_n \mathbf{f}_i^0  (\mathbf{f}_i^0)^T$ (note that $\mathbf{f}_i^0$ is already centered).
The covariance matrix of the transformed points $\mathbf{\hat{F}}^0_s$ will become the identity matrix: $cov(\mathbf{\hat{F}}^0_s) = \mathbf{I}_{c}$.

As the first part of the adaptation, the process is reversed by   rotating and scaling the whitened points using the \textit{SVD} on the target covariance matrix 
\[ 
\mathbf{\Sigma}_t = cov(\mathbf{F}^0_t) \in \mathbb{R}^{c \times c}
\]
\[ 
\mathbf{U}_t \mathbf{S}_t \mathbf{V}^*_t = svd(\mathbf{\Sigma}_t)
\]
\begin{equation}\label{recoloring}
\mathbf{F}^0_{s \rightarrow t} 
= \mathbf{\hat{F}}^0_s  \mathbf{S}^{\frac{1}{2}}_t (\mathbf{U}_t)^T,
\end{equation}
which yields the transformed features that now have the target covariance: $cov(\mathbf{F}^0_{s \rightarrow t} ) = \mathbf{\Sigma}_t$.
Secondly, we shift the transformed feature matrix by the target mean and reshape it back to the image format:
\[ 
\mathbf{F}_{s \rightarrow t}  = \mathbf{F}^0_{s \rightarrow t}  + \mathbf{\bar{f}}_t
\]
\begin{equation}\label{add_target_mean}
\mathbf{F}_{s \rightarrow t}  
\rightarrow 
\mathbf{x}^i_{s \rightarrow t} \in \mathbb{N}^{h_s \times w_s \times c}.
\end{equation}
Using \textit{FDM}, the transformed image has the mean and covariance of the target, while keeping the source content.
This process is illustrated in fig.~\ref{fig:FDM}. An example of the image transformation is shown in fig.~\ref{fig:da-configurations}.

Due to the generalized transformation, the same procedure can also be easily applied to the output of a certain layer $\varPhi_k(\mathbf{x})$ of a CNN, to use the higher level layer representation for the distribution alignment.
For this, the \textit{FDM} is applied to the layer responses of the source and target images and used as input for the following layer:
\begin{equation}
\varPhi_{k+1}(\mathbf{\mathbf{x}})= fdm(\varPhi_k(\mathbf{x}^i_s), \varPhi_k(\mathbf{x}^j_t)).
\end{equation}
In our experiments, we chose the first convolution layer before the non-linearity ($k=1,~c=64$)
and we refer to this method as \textit{FDM\_conv\_1}.

%% file: hist_matching.tex
\textit{Histogram Matching (HM)} (sometimes also called \textit{Histogram Specification}), is a common approach in image processing for finding a monotonic mapping between a pair of image histograms~\cite{Gonzalez2006}. It manipulates pixels of a source image in such a way that its histogram matches that of a target image. Given a source $\mathbf{x}_s^i \in \mathcal{D}_s$ and target $\mathbf{x}_t^j \in \mathcal{D}_t$ image, their histograms $h_{\mathbf{x}_s^i}$ and $h_{\mathbf{x}_t^j}$ are computed independently for all color channels. The corresponding cumulative distribution functions are obtained as

\begin{equation}
\begin{aligned}
cdf_{\mathbf{x}_s^i}(k) = \frac{1}{N_s} \sum\limits_{\ell=1}^k h_{\mathbf{x}_s^i}(\ell) \\
cdf_{\mathbf{x}_t^j}(k) = \frac{1}{N_t} \sum\limits_{\ell=1}^k h_{\mathbf{x}_t^j}(\ell),
\end{aligned}
\label{hm_cdf} \quad k = 1, \ldots, m
\end{equation}
where $m$ is the number of bins, $N_s$ and $N_t$ is the number of pixels in the source and target image, respectively. For every value $v$ in the source image the corresponding value $cdf_{\mathbf{x}_s^i}(v)$ is used to look up the value $v'$ so that both cumulative distribution functions are equal: $cdf_{\mathbf{x}_s^i}(v) = cdf_{\mathbf{x}_t^j}(v')$.
Therefore, \textit{HM} performs a mapping that optimally transforms intensities of the input image towards the target image. Fig.~\ref{fig:histogram-matching} demonstrates the procedure of applying \textit{HM} to a pair of images from different data sets.

%% file: experimental_results.tex
\begin{figure*}[!ht]
	\begin{center}
		\includegraphics[width=17.5cm]{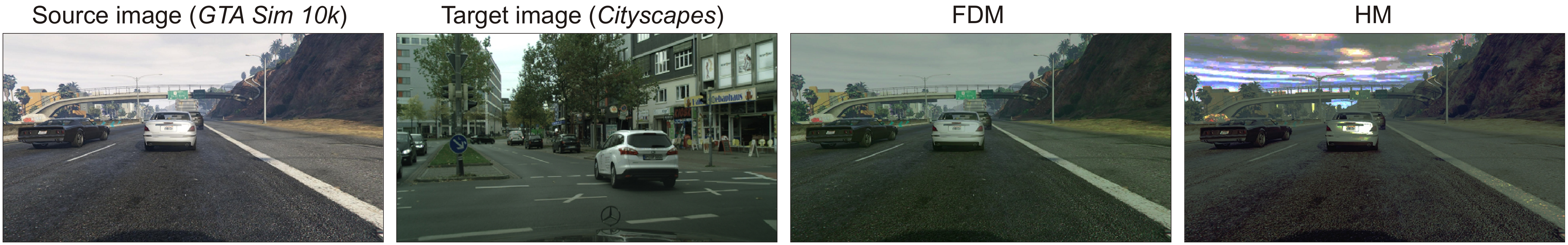}
		\caption{An exemplary DA procedure applying \textit{FDM} and \textit{HM} is shown. The leftmost image is from the \textit{GTA Sim 10k} data set and represents the source domain. The second image is taken from the target domain of the \textit{Cityscapes} data set and used to apply the \textit{DA} approaches. The two images on the right show the result of \textit{FDM} and \textit{HM}, respectively. It can be seen that the content of the initial image is preserved but the style and colors have been adapted. 
			\label{fig:da-configurations}}
	\end{center}
\end{figure*}

In this section we report on the experiments that were performed to assess and quantitatively measure the benefit of the proposed \textit{DA} approaches. As a performance measure the \textit{Faster R-CNN}~\cite{Ren2015} object detection model is trained and its accuracy is evaluated. The data used for training is given by images from the source domain and corresponding instance-level annotations: class labels and bounding boxes. In addition, unlabeled images from the target domain are used to apply the \textit{DA} methods. The trained models are evaluated on annotated images from the target domain. Unless stated otherwise we analyze the performance for detecting objects of the class \textit{car} due to the limited class diversity in some of the data sets and to make the experiments comparable. As a baseline for the comparison, the original \textit{Faster R-CNN} is trained on the source domain training data  and evaluated on the target domain test data without applying \textit{DA} methods.

We investigate the performance of the proposed \textit{DA} approaches \textit{FDM} and \textit{HM} described in section \ref{sec:fdm} for different domain shift scenarios. Fig.~\ref{fig:da-configurations} shows the domain adapted result obtained with the two methods applied on an exemplary pair of simulated source and real target image. In addition to the sole application of the approaches, the performance of their disjunctive (\textit{FDM} or \textit{HM}) and additive (applying first \textit{FDM} and then \textit{HM} with the same target domain image) combinations is evaluated. For one domain shift scenario we exemplary show the performance of the feature level adaptation method \textit{FDM\_conv\_1} described at the end of section \ref{sec:fdm}.

In all cases \textit{DA} is performed during the training process, which means that in each iteration of the training procedure, the \textit{DA} method is applied to the current training image using a randomly selected image of the target training data. To accelerate the training process for a specific domain shift scenario, generating a domain adapted training data set beforehand would be beneficial. In all experiments we initialize the network with a model pretrained for ImageNet classification. To fine-tune the detector we train 50k iterations with a learning rate of $10^{-3}$ followed by 40k iterations with a learning rate of $10^{-4}$. A momentum of $0.9$ and a weight decay of $5 \times 10^{-4}$ is used for training the networks in the presented experiments.

Trained models are evaluated adopting the \textit{PASCAL VOC} metric~\cite{Everingham2015}: the average precision (AP) of the class \textit{car} is computed with an intersection over union (IoU) threshold of 0.5 for positive detections. There are several steps in the experimental procedure that make use of random number generation, e.g. weight initialization and selection of target training images for \textit{DA}. This can lead to a serious spread of the results when repeating experiments. Note that this cannot be solved by a fixed random seed due to the different input data~\footnote{CACE principle: Changing Anything Changes Everything~\cite{sculley2015hidden}.}. To estimate the statistical uncertainty of the result and to obtain a meaningful assessment of the benefit of the proposed approaches, the procedure is repeated at least six times for each configuration. For a comprehensive comparison, we report, additionally to the maximum AP, the mean and standard deviation for each experimental configuration.

In the following sections the results of the experiments for different domain shift scenarios are discussed. In the experiments reported in the first two sections, synthetic data is utilized as source domain. While the target domain in section \ref{sec:synthetic_data_enhancement} is given by the \textit{KITTI} data set, section \ref{sec:learning_from_synthetic_data} documents the domain shift to the \textit{Cityscapes} data set as target domain. In section \ref{sec:foggycity} a domain shift from \textit{Cityscapes} to \textit{Foggy Cityscapes}, which is a synthetically modified version of the original \textit{Cityscapes}, is performed. The last two sections concern the domain shift between the real data sets \textit{Cityscapes} and \textit{KITTI} in both directions.

\input{tab_gta_2_kitti.tex}

\subsection{From Los Santos to Karlsruhe}
\label{sec:synthetic_data_enhancement}
The computing power of modern GPUs can be used to train neural networks, but their original purpose is the generation of computer graphics for e.g. video games. The progress in computer graphics allow for  creation of more and more realistic synthetic data and as the effort to additionally generate annotations for this data is minimal, it seems likely to use it for the training of \textit{CNNs}. However, applying a network trained on simulated data to real images does not reach the performance of a network directly trained on target domain data. Making use of \textit{DA} methods, reduces this deficit and thus increases the usability of simulation for the training of \textit{CNNs} on synthetic data. In this section we evaluate the proposed methods for the domain shift from the synthetic \textit{GTA Sim 10k} data set and the real world images of the \textit{KITTI} data set.

The \textit{GTA Sim 10k} data set consists of $10000$ simulated images of the fictional city Los Santos generated with the video engine of the game Grand Theft Auto V (GTA V) and annotations in form of object categories and bounding boxes. As target domain we utilize images from the \textit{KITTI} 2D object detection data set \footnote{\url{http://www.cvlibs.net/datasets/kitti/eval_object.php}} which was recorded in the city of Karlsruhe. The $7518$ test images are used as target domain data for applying the \textit{DA} methods. The $7481$ training images are used as test data for evaluation as only for these images labels are publicly available. To show that our \textit{DA} methods not only improve the performance for easy to detect objects, the object detection performance is analyzed with respect to all three difficulty categories easy, moderate and hard, which are provided for the \textit{KITTI} data set.

Table \ref{tab:gta_2_kitti} shows the results of the \textit{Faster R-CNN} performance when using the \textit{GTA Sim 10k} data for training with and without applying \textit{DA} to the data. In addition, we show the result of training the \textit{CNN} with the five times larger \textit{GTA Sim 50k} (50000 simulated images) data set for comparison. Applying \textit{FDM} or \textit{HM} methods for the domain shift results in a significant boost compared to using the plain \textit{GTA Sim 10k} data set. It even performs on the same level as the model obtained when training with the \textit{GTA Sim 50k} data set without any modification. This experiment shows that using our proposed \textit{DA} methods greatly increases the value of smaller simulated data sets.

\subsection{From GTA Sim 10k to Cityscapes}
\label{sec:learning_from_synthetic_data}
In this section we examine a different domain shift from simulated data to real world images. The \textit{GTA Sim 10k} data set again represents the source domain. For this experiment the target domain is given by \textit{Cityscapes} data set. This data set was recorded with a car equipped with a camera and covers urban scenarios from several German cities, Strasbourg and Zurich. As ground truth it provides instance-level pixel-wise annotations for eight categories. Bounding boxes for object detection are not included in the annotations, but they can be computed as axis-aligned minimum bounding rectangles from each instance contour. This was done using the data preparation script \footnote{\url{https://github.com/yuhuayc/da-faster-rcnn/prepare_data/prepare_data.m}} provided by Chen et al.~\cite{Chen2018}. The training set contains $2975$ images which are used as unlabeled target domain data for \textit{DA}. The object detection performance is evaluated based on the $500$ images of the validation set and corresponding annotations.

For this experiment, the variations of the approaches proposed in section \ref{chap:methods} have been tested and the results are compared to the plain \textit{Faster R-CNN} (without \textit{DA}) and the methods described in \textit{DA Faster R-CNN}~\cite{Chen2018}, \textit{Strong-Weak DA}~\cite{Saito2019}, and \textit{CycleGAN}~\cite{Zhu2017}~\footnote{We trained the \textit{CycleGAN} for $50$ epochs on all training images from both domains to translate the \textit{GTA Sim 10k} data set into the \textit{Cityscapes} domain. After that, the \textit{Faster R-CNN} model was trained on the translated images.}. Note that for the same method and experimental setup different values have been reported by different publications, despite source code and data sets being publicly available. The results presented in table \ref{tab:gta_2_city} show that all variations of \textit{FDM} and \textit{HM} perform comparable or better than the method used in~\cite{Chen2018}. While the overall best result is obtained with the method presented in \textit{Strong-Weak DA}~\cite{Saito2019}, we still achieve a good result utilizing \textit{FDM\_conv\_1}, which has a much simpler training procedure. One can also conclude from the table that combining \textit{FDM} and \textit{HM}  is a valid procedure as it performs better than applying each method separately.

\input{tab_gta_2_city.tex}

\subsection{Driving in the Fog}
\label{sec:foggycity}
In the next section we evaluate our proposed approaches in the domain shift scenario from \textit{Cityscapes} as source domain to \textit{Foggy Cityscapes} as target domain. \textit{Foggy Cityscapes} is a synthetic data set that was created simulating fog on the real scenes of \textit{Cityscapes}. The annotations are the same as for the original data set, i.e., the bounding boxes correspond to the tightest rectangle around instance contours. We use the designated parts of both data sets for training and the $500$ validation images of \textit{Foggy Cityscapes} for measuring performance.

Since both data sets contain the same class categories, mean average precision (mAP) over the eight classes is computed to evaluate the object detection performance. The result for the different \textit{DA} methods is shown in table \ref{tab:city_2_foggy_city}. The combination of \textit{FDM} and \textit{HM} yields the best result of our proposed approaches and the maximum achieved mAP is comparable to the one using the method published in~\cite{Saito2019}. Again performing \textit{FDM} and \textit{HM} after each other yields an additional boost compared to the individual application of $~4\%$.

\input{tab_city_2_foggy_city.tex}
\input{tab_kitti_2_city.tex}
\input{tab_city_2_kitti.tex}
\subsection{Domain Adaptation between KITTI and Cityscapes}
\label{sec:crosscamera}
The last experimental scenarios of this work deal with the domain shift between two real image data sets. We evaluate the proposed \textit{DA} approaches using the \textit{KITTI} data set as source domain and the \textit{Cityscapes} data set as target domain and vice versa. These two data sets both contain real images but have been captured with different sensor setups and therefore differ in image styles and camera position. In addition, they were recorded in different locations and lighting and weather conditions and can therefore be regarded as two distinct domains. The results are shown in table \ref{tab:kitti_2_city} for the domain shift from \textit{KITTI} to \textit{Cityscapes} and in table \ref{tab:city_2_kitti} for the opposite direction. These two experiments were also performed in~\cite{Chen2018} and the corresponding results have been added to the tables for comparison.

We have evaluated the \textit{FDM} and \textit{HM} methods and their disjunctive and additive application. For the domain shift to \textit{Cityscapes}, all variations are on the same level and achieve an AP for the class \textit{car} of approximately $3\%$ more compared to \textit{DA Faster R-CNN}~\cite{Chen2018} and $6-7\%$ more than the baseline without \textit{DA}. A similar picture can be observed for the opposite domain shift (table \ref{tab:city_2_kitti}) where the best result of $66.93\%$ is obtained using the variation that uses \textit{FDM} or \textit{HM}. The AP is again nearly $3\%$ higher than \textit{DA Faster R-CNN}~\cite{Chen2018}. Note that in all scenarios a combination of \textit{FDM} and \textit{HM} is better than the single application of one of them.

%% file: tab_gta_2_kitti.tex
\begin{table*} [!ht]
	\begin{center}

		\begin{tabular}{ccccc|cccc|cccc}
			\midrule
			Method & \multicolumn{2}{c}{Max AP \textit{Easy}} & Mean & Std & \multicolumn{2}{c}{Max AP \textit{Mod.}}  & Mean & Std & \multicolumn{2}{c}{Max AP \textit{Hard}} & Mean & Std \\ 
 			\midrule
 			
			Faster R-CNN 10k & 61.61 &  & 58.64 & 1.37 & 43.25 & & 42.24 & 0.68 & 35.08 &  & 34.16 & 0.62 \\ 

			& 55.42 & \cite{Roberson2017} & - & - & 38.28 & \cite{Roberson2017} & - & - & 29.04 & \cite{Roberson2017} & - & - \\

			Faster R-CNN 50k & 66.59 & & 65.19 & 1.23 & 50.13 & & 49.31 & 0.80 & 38.49 & & 37.87 & 0.46     \\ 
			& 68.56 & \cite{Roberson2017} & - & - & 50.08 & \cite{Roberson2017} & - & - & 39.26 & \cite{Roberson2017} & - & - \\ 
			\midrule
	
			FDM Sim 10K & \textbf{68.72} &  & 68.13 & 0.44 & \textbf{50.92} &  & 49.67 & 1.91 & \textbf{39.85} &  & 39.09 & 0.92  \\ 
			HM Sim 10K & 68.38 &  & 67.45 & 0.70 & 49.81 & & 48.40 & 1.59 & 38.91 &  &  38.45 & 0.49 \\ 
			\midrule
		\end{tabular} 
		\caption{Average precision for the class \textit{car} of object detectors trained on synthetic \textit{GTA Sim} data evaluated on real images from the \textit{KITTI} data set. The first four rows show the result for the baseline \textit{Faster R-CNN} model for $10k$ and $50k$ training images. The two lower rows show the result for the training with $10k$ images and the proposed \textit{DA} methods. For all models trained by us, the training schedule described in section \ref{chap:results} was used, and we report mean and standard deviation in addition to the maximum achieved AP. Note that a different training schedule was used for this domain shift scenario in \textit{Driving in the Matrix}~\cite{Roberson2017}. The \textit{FDM} method yields the best performance of the detectors trained on $10k$ images and is even on par with the plain \textit{Faster R-CNN} trained on five times as many images.}
		\label{tab:gta_2_kitti}
	\end{center}
\end{table*}

%

%

%% file: tab_gta_2_city.tex
\begin{table} 
	\begin{center}
		
		\begin{tabular}{ccccc}	
			\midrule
			
			Method & \multicolumn{2}{c}{Max AP Car} & Mean & Std  \\ 
			\midrule
			
			Faster R-CNN & 33.69 & & 33.46 & 0.25  \\ 
			& 30.12 & \cite{Chen2018} & - & -  \\
			& 34.60 & \cite{Saito2019} & - & -  \\
			
			DA Faster R-CNN & 38.97 & \cite{Chen2018} & - & -  \\ 
			& 34.20 & \cite{Saito2019} & - & -  \\ 
			Strong-Weak DA & 42.30 & \cite{Saito2019} & - & -  \\
			CycleGAN & 39.79 & & 39.51 & 0.25  \\ 
		    \midrule
			FDM & 38.45 & & 38.28 & 0.16  \\ 
			HM & 39.34 & & 38.24 & 0.80  \\ 
			FDM or HM & 39.78 & & 38.74 & 0.59  \\ 
			FDM and HM & 40.05 & & 39.39 & 0.47  \\ 
			FDM\_conv\_1 & \textbf{40.55} & & 40.40 & 0.12  \\

			\midrule
		
		\end{tabular} 
		\caption{Quantitative analysis results of the \textit{DA} from \textit{GTA Sim 10k} to the \textit{Cityscapes} data set showing maximum AP for the class \textit{car}. The best result is achieved by the method proposed in \textit{String-Weak DA} including tuning of hyper-parameters for this specific scenario. The \textit{FDM\_conv\_1} gives the best AP of the methods presented in this work and the second best overall performance. Combining \textit{FDM} and \textit{HM} yields higher AP then each used independently. }
		\label{tab:gta_2_city}
	\end{center}
\end{table}

%

%% file: tab_city_2_foggy_city.tex
\begin{table}
	\begin{center}

			\begin{tabular}{ccccc}
				\midrule
				Method & \multicolumn{2}{c}{Max mAP} & Mean & Std  \\ 
				\midrule
				Faster R-CNN & 18.84 & & 18.46 & 0.29   \\ 
				 & 18.80 & \cite{Chen2018} & - & -  \\ 
				 & 20.30 & \cite{Saito2019} & - & -  \\

				DA Faster R-CNN & 27.6 & \cite{Chen2018} & - & -  \\ 
				 & 22.5 & \cite{Saito2019} & - & -  \\
	
				Strong-Weak DA & 34.3 & \cite{Saito2019} & - & -  \\ 
	
				\midrule
				FDM & 29.08 & & 27.92 & 0.76  \\ 
				HM & 29.24 & & 28.04 & 0.80  \\ 
				FDM or HM &  30.81 & & 30.11 & 0.39  \\ 
				FDM and HM & \textbf{34.00} & & 32.96 & 0.89  \\ 
				\midrule
			\end{tabular} 
		\caption{Results for adapting \textit{Cityscapes} to \textit{Foggy Cityscapes}. For this domain shift scenario the mAP over all classes that are provided with instance-level annotations in the data sets are shown. The additive combination of \textit{FDM} and \textit{HM} is on par with the best result that was obtained by \textit{Strong-Weak DA}.}
		\label{tab:city_2_foggy_city}
	\end{center}
\end{table}

%% file: tab_kitti_2_city.tex
\begin{table}
	\begin{center}

		\begin{tabular}{ccccc}
            \midrule
			Method & \multicolumn{2}{c}{Max AP Car} & Mean & Std \\ 
            \midrule
			Faster R-CNN & 34.97 & & 34.85 & 0.12 \\ 
			 & 30.20 &\cite{Chen2018} & - & -  \\
			DA Faster R-CNN & 38.50 & \cite{Chen2018} & - & -  \\ 
			\midrule
			FDM & 41.52 & & 41.09 & 0.43  \\ 
			HM & 41.18 & & 40.50 & 0.53  \\ 
			FDM or HM & \textbf{41.79} & & 41.13 & 0.42  \\ 
			FDM and HM & 41.22 & & 40.78 & 0.38 \\ 
            \midrule
		\end{tabular}
		\caption{Quantitative results of the domain transfer from \textit{KITTI} to \textit{Cityscapes} showing AP for class \textit{car}. Compared to \textit{DA Faster R-CNN} all of the evaluated \textit{FDM} and \textit{HM} variations show a boost of approximately $3\%$. The best performance is obtained by the disjunctive combination.}
		\label{tab:kitti_2_city}
	\end{center}
\end{table}

%% file: tab_city_2_kitti.tex
\begin{table}
	\begin{center}
		
		\begin{tabular}{ccccc}
			\midrule
			Method & \multicolumn{2}{c}{Max AP Car} & Mean & Std  \\ 
			\midrule
			Faster R-CNN & 64.61 & & 64.14 & 0.45  \\ 
			Faster R-CNN & 53.50 &\cite{Chen2018} & - & -  \\ 
			DA Faster R-CNN & 64.10 & \cite{Chen2018} & - & -  \\ 
			\midrule
			FDM & 66.52 & & 65.87 & 0.42  \\ 
			HM & 66.23 & & 66.00 & 0.22 \\ 
			FDM or HM & \textbf{66.93} & & 66.25 & 0.41 \\ 
			FDM and HM & 66.15 & & 65.87 & 0.17  \\
			\midrule
		\end{tabular} 
	    \caption{AP for the class \textit{car} in the domain shift scenario \textit{Cityscapes} to \textit{KITTI}. All of the proposed DA variations perform $2-3\%$ better than \textit{Faster R-CNN} and \textit{DA Faster R-CNN}. The best result is again achieved with the disjunctive combination of \textit{FDM} and \textit{HM}.}
		\label{tab:city_2_kitti}
	\end{center}
\end{table}

%

%% file: conclusion.tex
In this paper, we suggest to use two approaches based on the alignment of global image statistics for \textit{unsupervised DA}. Applying the described methods to train a \textit{Faster R-CNN} yields state-of-the-art object detection performance in the presence of a shift between domains without additional labeling effort. The performance is validated in different domain shift scenarios and compared to other recently published approaches. The scenarios include the transfer from simulated to real data sets as well as usage of training data from one camera to detect objects on images from another camera. Compared to \textit{DA Faster R-CNN} and \textit{Strong-Weak DA}, no architectural modifications to the detector model or tuning of hyper-parameters of the training procedure for every particular domain shift are necessary, since the approaches only alter the training input images. This makes the methods model-invariant and easily applicable to other object detection models, such as \textit{SSD}~\cite{Liu2016} or \text{YOLO}~\cite{Redmon2017}, and allows to apply them for other tasks, e.g. semantic segmentation. Another consequence is that the time needed for training the model is short compared to more complicated architectures and approaches, especially when generating a data base with modified images in advance. We also show that applying the methods for the training on a synthetic data set is on par with using five times as much synthetic data without \textit{DA}. This shows that for the training of neural networks the quality of the data is more important than its quantity. As future work we plan to investigate combinations with other \textit{DA} approaches and to apply our techniques to higher network layers.